\documentclass[]{interact}

\usepackage{epstopdf}% To incorporate .eps illustrations using PDFLaTeX, etc.
\usepackage[caption=false]{subfig}% Support for small, `sub' figures and tables

\usepackage[numbers,sort&compress]{natbib}% Citation support using natbib.sty
\bibpunct[, ]{[}{]}{,}{n}{,}{,}% Citation support using natbib.sty
\theoremstyle{plain}% Theorem-like structures provided by amsthm.sty

\theoremstyle{definition}

\theoremstyle{remark}

\usepackage{array}
\usepackage{tabularx}
\usepackage{comment}
\usepackage{rotating}
\usepackage{lipsum}% http://ctan.org/pkg/lipsum
\usepackage{multicol}
\usepackage{placeins}
\usepackage{adjustbox} %smn
\usepackage{makecell} %smn
\usepackage{arydshln} %smn
\usepackage{latexsym}
\usepackage{times}
\usepackage{latexsym}
\usepackage{url}
\usepackage{hyperref}
\usepackage{multirow}
\usepackage{float}
\usepackage{subfig}
\usepackage[]{xcolor}
\usepackage{diagbox}
\usepackage{makecell}
\usepackage{tikz}
\usepackage{array}
\usepackage{tabularx}
\usepackage{graphicx}
\usepackage{color,array}
\usepackage{comment}
\usepackage{rotating}
\usepackage{lipsum}% http://ctan.org/pkg/lipsum
\usepackage{multicol}
\usepackage{placeins}
\usepackage{adjustbox} %smn
\usepackage{makecell} %smn
\usepackage{arydshln} %smn
\definecolor{backgG}{RGB}{255, 255, 153}
\definecolor{tagtxtG}{RGB}{102, 102, 0}
\definecolor{backgPc}{RGB}{179, 255, 179}
\definecolor{tagtxtPc}{RGB}{0, 102, 0}
\definecolor{backgPw}{RGB}{255, 179, 179}
\definecolor{backgPw}{rgb}{0.0, 1.0, 1.0}
\definecolor{tagtxtPw}{RGB}{0.0, 1.0, 1.0}
\definecolor{backgPo}{rgb}{0.76, 1, 1}
\definecolor{tagtxtPo}{RGB}{102, 0, 0}
\definecolor{backgPm}{rgb}{0.98, 0.81, 0.69}
\definecolor{tagtxtPm}{RGB}{0,1,1}

\newcommand{\orcid}[1]{\href{https://orcid.org/#1}{\includegraphics[width=8pt]{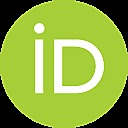}}}

\begin{document}

\title{Hallucination Reduction in Long Input Text Summarization}
%\begin{comment}  
\author{
\name{Tohida Rehman \orcid{0000-0002-3578-1316} \textsuperscript{a}, 
Ronit Mandal \textsuperscript{a},
Abhishek Agarwal \textsuperscript{a}, 
Debarshi Kumar Sanyal \orcid{0000-0001-8723-5002} \textsuperscript{b}}
\affil{\textsuperscript{a} Department of Information Technology, Jadavpur University, Kolkata -- 700106, India. (tohidarehman.it@jadavpuruniversity.in; ronitmandal21@gmail.com; abhi2001agarwal@gmail.com) \\
\textsuperscript{b} School of Mathematical and Computational Sciences, Indian Association for the Cultivation of Science, Kolkata -- 700032, India. (debarshi.sanyal@iacs.res.in)}
}
\maketitle

\begin{abstract}
Hallucination in text summarization refers to the phenomenon where the model generates information that is not supported by the input source document. Hallucination poses significant obstacles to the accuracy and reliability of the generated summaries. In this paper, we aim to reduce hallucinated outputs or hallucinations in summaries of long-form text documents. We have used the PubMed dataset, which contains long scientific research documents and their abstracts. We have incorporated the techniques of data filtering and joint entity and summary generation (JAENS) in the fine-tuning of the Longformer Encoder-Decoder (LED) model to minimize hallucinations and thereby improve the quality of the generated summary. We have used the following metrics to measure factual consistency at the entity level: precision-source, and F1-target. Our experiments show that the fine-tuned LED model performs well in generating the paper abstract. Data filtering techniques based on some preprocessing steps reduce entity-level hallucinations in the generated summaries in terms of some of the factual consistency metrics. 
%\textcolor{blue}{Data filtering techniques based on some preprocessing steps reduce entity-level hallucinations in the generated summaries in terms of some of the factual consistency metrics. However, the modified training methods do show a reduction in ROUGE, METEOR, and BERTScore values.}   
\end{abstract}


\section{Introduction}
\label{intro}
With the exponential growth of textual data, the need for effective summarization techniques becomes crucial to extracting relevant and concise information from lengthy documents. Text summarization plays a vital role in various domains, including news articles, legal documents, and scientific papers. However, when it comes to handling long input texts, such as research papers or legal documents, the task becomes even more challenging. The input documents of such tasks are often significantly longer than the maximum context lengths of most standard transformer models. This has motivated researchers to explore changes in model architecture and training strategies. For instance, to avoid the quadratic growth in memory consumption of the attention computation in transformers, many memory-efficient transformer variants have been proposed in \cite{huang2021efficient}. To handle long inputs, Beltagy et al. \cite{beltagy2020longformer} have added a long input pre-training stage to the transformer while Chalkidis et al. \cite{Chalkidis2022ldc} have only fine-tuned the models with long inputs without any pre-adaptation.

In the context of long input text summarization, one common issue is the presence of \textit{hallucinations}, that is, the generated summary includes factual inconsistencies or introduces information not present in the source document. Hallucinations can be categorized as intrinsic and extrinsic hallucinations \cite{maynez-etal-2020-faithfulness}. Intrinsic hallucinations occur when the model interprets information from the input text incorrectly but uses the terms or concepts that occur in the source document. Extrinsic hallucinations occur when the model generates text that does not match the input text, that is, uses terms and concepts not even present in the source document. Hallucinations can undermine the reliability and accuracy of the summarization process, potentially leading to misinformation or misleading interpretations. These contradictions in fact can exist at the entity or phrase level. A model-generated summary may include named-entities that were not included in the source document. This is known as the entity hallucination problem \cite{nan-etal-2021-entity}. 

The main contributions of this paper are:
\begin{enumerate}
     \item We use Longformer Encoder-Decoder (LED) model \cite{beltagy2020longformer} to generate summary of scientific articles in the PubMed dataset \cite{cohan-etal-2018-discourse}. In addition, we explore two techniques, namely, data filtering and JAENS (Join sAlient ENtity and Summary generation) \cite{nan-etal-2021-entity} to study their effect on the factual consistency of the generated summaries.
     \item We analyze the factual consistency of the output summary at the entity level using the following metrics: precision-source and F1-target, introduced by \cite{nan-etal-2021-entity}. We also use the traditional metrics, namely, ROUGH \cite{lin2004rouge}, METEOR \cite{banerjee2005meteor}, and BERTScore \cite{zhang2019bertscore}, to evaluate the performance of the models. The entity-based data filtering technique improves the precision-source but the other metrics achieve higher values when fine-tuning with LED is done without the other two techniques. 
     Our code and results are available on github \footnote{\url{https://github.com/tohidarehman/Hallucination-Reduction-Text-Summarization}}.
\end{enumerate}

\section{Literature survey}
Early research efforts in text summarization predominantly focused on extractive methods, which involve selecting the most significant sentences or phrases from the source document to form the gist. 
While extractive summarization approaches achieved reasonable success, it was hard to modify these methods to handle information that required rephrasing or merging content from multiple sentences. This limitation led to research in abstractive summarization techniques, which aim to generate summaries by understanding the source text and producing new sentences that capture the essential information. 
The emergence of recurrent neural networks (RNNs) that are capable of processing and producing text has significantly improved abstractive summarization, but they sometimes exhibit undesirable behavior such as incorrectly reproducing factual details, an inability to deal with out-of-vocabulary (OOV) words, and repetitive themselves \cite{nallapati-etal-2016-abstractive}.
The pointer-generator model with a coverage mechanism helps to resolve the problem of out-of-vocabulary (OOV) words, and repeating phrase generation \cite{See2017GetTT,rehman2021automatic,rehman-etal-2022-named,rehman2023research, 10172215}.

Large pre-trained transformer models have proven to be exceptionally capable of dealing with natural language tasks \cite{devlin2018bert,10.1007/978-981-19-1657-1_21}. Handling extended textual sequences, on the other hand, remains a considerable issue for these models. These challenging input documents are often substantially longer than the maximal context lengths of typical transformer models, necessitating both specialized model architectural adjustments and unique training regimes to accommodate. For example, numerous memory-efficient transformer variations have been proposed to prevent the quadratic escalation in memory consumption of the attention estimation in transformers. Another severe issue is the inability of current abstractive summarization methods to generate faithful results. 
These systems frequently struggle to verify that the generated summaries only include information extracted from the source document and do not include manufactured or hallucinated statements. These hallucinations can occur for a variety of causes, including biases in the training data, a lack of context perception, or model over-optimization. 
Cao et al. \cite{CaoWeiLiLi2018} and Kryściński et al. \cite{kryscinski-etal-2020-evaluating} reported that approximately 30\% of the summaries generated by seq2seq models suffer from the issue of hallucination. As a result, as noted in the works of the NLP community, attention has been drawn more and more to the faithfulness and factual components of abstractive summarization \cite{kryscinski-etal-2020-evaluating, goyal-durrett-2020-evaluating, zhu-etal-2021-enhancing}. 
Many recent works study entity-level and relation-level hallucination problems in the generated text. Nan et al. \cite{nan-etal-2021-entity} address entity hallucination by applying a filter on the training data and multi-task learning. 
Goyal and Durrett \cite{goyal-durrett-2020-evaluating} study relation hallucination, that is, whether the semantic relationships manifested by the individual dependency arcs in a generated sentence are entailed by the source sentence.
One notable work by Narayan et al. \cite{narayan-etal-2018-dont} incorporates entity chain content planning to guide faithful summary generation. There has been growing interest in quantitatively measuring the faithfulness of text generation models.  Most widely-adopted evaluation metrics for text generation, such as ROUGE \cite{lin2004rouge} and BERTScore \cite{zhang2019bertscore} correlate poorly with the human perceived faithfulness of the generated text \cite{kryscinski-etal-2020-evaluating}. Recent studies explore categorical and content-based analysis for measuring the faithfulness of summaries \cite{goyal-durrett-2020-evaluating}. 

\section{Methodology}
To handle long input sequences, we utilized the pre-trained checkpoints of the Longformer Encoder Decoder (LED) model \cite{beltagy2020longformer}, which incorporates a sliding window and dilated sliding window attention mechanisms. It consists of both the encoder and decoder Transformer stacks, but instead of using full self-attention in the encoder, it employs the Longformer's efficient local+global attention pattern. The decoder applies full self-attention to all encoded tokens and previously decoded locations. Because pre-training LED is expensive, authors in \cite{beltagy2020longformer} have used BART parameters to initialize LED parameters and adhered to BART's exact design in terms of the number of hidden sizes and layers.
This allows it to effectively process lengthy inputs. We performed fine-tuning of the pre-trained LED model to adapt it specifically for text summarization of scientific documents. To ensure the accuracy of the summaries, we implemented scispaCy-based Named Entity Recognition (NER) on the ground truth summaries. We applied the JAENS (Jointly Aligned Entity Names and Summaries) approach to augment salient entities in front of the abstracts. Training the model to recognize summary-worthy named-entities aims to enhance the precision and recall related to named-entities in the generated summaries.

We have performed experiments with 3 variants with the LED model: (1) fine-tuned on the LED model, (2) fine-tuned LED model with the filtered dataset, and (3) fine-tuned LED model using the JAENS approach on the filtered dataset.

\subsection{Fine-tuning LED}
Pre-trained models like LED learn rich language representations from a large corpus. Fine-tuning customizes these models for specific tasks. It initializes the model with pre-trained weights, then fine-tunes it on a task-specific dataset using backpropagation. Fine-tuning leverages the model's language understanding saves time and resources, and requires less labeled data. This approach enhances text summarization by adapting the model to task-specific data while leveraging its pre-trained knowledge.

\subsection{Entity-based data filtering}
As demonstrated successfully by \cite{nan-etal-2021-entity}, the training dataset's quality has a significant impact on the amount of entity-level hallucinations present in the generated summary. With that in mind, we applied scispaCy  Named Entity Recognition (NER) to the gold summary for the PubMed dataset. This allows us to identify all the named-entities present in the gold summary. Our objective is to ensure that these named-entities have corresponding $n$-gram matches within the source document. For unigram matching, we avoid matching any stop words. Therefore, if any named-entity of a sentence in the summary cannot be found within the source document, we decided to exclude that sentence from the summary. If the number of sentences in the summary is one and using the filtering technique we need to remove that sentence, then the entire article-summary pair has been removed from the dataset.

\subsection{Joint sAlient ENtity and Summary generation (JAENS)}
The JAENS (Joint sAlient ENtity and Summary generation) approach, originally introduced by Nan et al. \cite{nan-etal-2021-entity}, is an alternative generative approach aimed at enhancing entity-level precision, and recall metrics. JAENS trains the LED model to construct a sequence that contains summary-worthy named-entities, a special token, and the summary itself, as opposed to typical summarization approaches. This approach enables the model to simultaneously learn the identification of summary-worthy named-entities while generating summaries, similar to the multitask learning approach. By prioritizing the generation of salient named-entities in the decoder, JAENS ensures that the summaries incorporate and highlight these important entities through decoder self-attention. By incorporating the JAENS approach into our project, we aim to mitigate entity-level summary hallucinations and improve the overall quality of the generated summaries.
\section{Experimental setup}
\subsection{Datasets}
We used a dataset collected from a scientific repository, PubMed\footnote{\url{https://pubmed.ncbi.nlm.nih.gov/}}), and was introduced in \cite{cohan-etal-2018-discourse}. We chose scientific papers as our dataset because they are examples of long documents with a standard discourse structure. Furthermore, scientific papers are rich in domain-specific terminology and technical information, which makes them an important source of information for researchers and practitioners alike.
PubMed is a biomedical literature database that contains over 30 million citations and abstracts of research articles.  The dataset contains almost 19,000 scholarly publications on diabetes from the PubMed database, which are categorized into one of three categories.
%Every publication is encoded by a TF-IDF-weighted word vector from a vocabulary of 500 distinct words. Additionally, the citation network of the PubMed dataset consists of 200000 samples. 
In our experiment, we choose, for training 2000 examples, validation 250 examples, and testing 250 examples. The size of the used dataset after applying the entity-based filtering procedure was 1798 examples for training, 232 examples for validation, and 236 examples for testing. The average number of sentences in summary before applying the entity-based data filtering technique was 7.33, 7.04, and 7.51 for training, validation, and test datasets. 
The average number of sentences in a summary after applying the entity-based data filtering technique is 4.34, 4.11, and 4.58 for training, validation, and test datasets.
\subsection{Data processing}
We eliminated all punctuation, numerals, special characters, mathematical formulas, and citation markers from the documents and lowercase the entire corpus. When we were going through documents, we made sure they were the right length and had the right structure. If something was too long, like a thesis, or too short, like a tutorial announcement, we removed it. We also looked for documents that did not have an abstract or a clear structure. To understand the structure, we used the section headings as clues. Sometimes, documents had figures or tables that did not help us understand the text. We got rid of those, keeping only the words.
%We also rem any math  into special codes so we could analyze them better. 
%In our used dataset, the average source document length is 18181 tokens, while the average gold summary length is 1288 tokens. 
In our model, the maximum number of allowed input tokens is 8192, that of output tokens is 512, and the minimum number of output tokens is 100 only.  
In line with the JAENS approach, we used the scispaCy model \texttt{en\_core\_sci\_sm} \footnote{\url{https://allenai.github.io/scispacy/}} library to generate summary-worthy named-entities and augmented the list of comma-separated named-entities before the ground truth summary (abstract) for each sample of the dataset. The sequence of named-entities is followed by a special token, which helps separate the entities from the abstract. This special token is chosen from the model's vocabulary such that it is not commonly occurring and can help the model learn to recognize the named-entities separately from the actual abstract. This helps in training the model as now the model will apply special attention to these entities while generating the summary.
\subsection{Implementation details}
We conducted our experiments using Google Colab Pro+, which provided us with an NVIDIA A100 GPU. For all experiments, we used the base variant of the pre-trained LED model \texttt{led-base-16384}\footnote{\url{https://huggingface.co/allenai/led-base-16384}}, due to resource limitations. 
Firstly, we fine-tuned the LED model on the original 2000-sample PubMed dataset. Secondly, we utilized a filtered version of the taken dataset by removing article-abstract pairs with a $prec_s$ score (to be defined in the next sub-section) less than 1 (i.e., we ensure that the abstract -- which is the ground-truth summary -- contains almost no hallucinations of entities) and performed fine-tuning on this filtered dataset. Finally, we incorporated the JAENS approach into the fine-tuning process by augmenting summary-worthy named-entities in front of the abstract for each example of the filtered train dataset, aiming to enhance entity-level precision, recall, and F1 metrics in the generated summaries and thus reduce the entity-level hallucinations. For all the models, we fine-tuned up to 10 epochs. To evaluate the models, we used the same test dataset that was obtained after the entity-based data filtering technique.

\subsection{Evaluation metrics}
We employ a comprehensive set of widely used text summarization evaluation metrics, including ROUGE \cite{lin2004rouge}, METEOR \cite{banerjee2005meteor}, BERTScore \cite{zhang2019bertscore}, to assess the quality and effectiveness of the generated summaries. Unfortunately, these metrics are inadequate to quantify factual consistency \cite{kryscinski-etal-2020-evaluating}. Hence, we have also used three new metrics, introduced by \cite{nan-etal-2021-entity}, to evaluate the factual consistency of the generated summaries. 

We define $\mathcal{N}(t)$ as the count of named-entities in the target (ground truth or gold summary) and $\mathcal{N}(h)$ as the count of named-entities in the hypothesis (generated summary). To determine the number of entities in the hypothesis that have corresponding matches in the source document, we use $\mathcal{N}(h \cap s)$. In circumstances when a named-entity in the summary spans many words, we consider it a match if any component of the named-entity can be identified in the original document, permitting partial matching based on $n$-grams. 
\textbf{Precision-source}, defined as, 
%\begin{align}
 $prec_s= \mathcal{N}(h \cap s) / \mathcal{N}(h) $ 
%\end{align}
is a metric that is used to determine the intensity of hallucination in relation to the source. Note that 
$prec_s$ represents the percentage of entities mentioned in the generated summary that can be retrieved from the source.  Low $prec_s$ indicates that hallucination is possibly present in the generated text. However, $prec_s$ does not capture the computed summary's entity-level correctness in relation to the ground-truth summary. Entity-level accuracy of the generated summary is calculated using the \textbf{precision-target} as
$prec_t= \mathcal{N}(h \cap t) / \mathcal{N}(h);$    
the \textbf{recall-target} as 
$recall_t = \mathcal{N}(h \cap t) / \mathcal{N}(t);$ and \textbf{F1 score} as 
$F1_{t} = \frac{2*(recall_t*prec_t)}{recall_t+prec_t}.
$ 
Here, $\mathcal{N}(h \cap t)$ represents the number of matched named-entities in the generated summary and the ground truth summary. 
%The \textbf{F1 score ($F1_t$)} is calculated as 
%\begin{align}
 %   F1_{t} = \frac{2*(recall_t*prec_t)}{recall_t+prec_t}
%\end{align}

Note that the above precision and recall scores can be calculated in two ways. One is to consider the entity mentioned in each document (which may be the source $s$ or target $t$ or hypothesis $h$) as a set so that multiple occurrences of an entity in a document are equivalent to a single occurrence. The other is to consider the entity mentioned in a document as a list; here, if a metric is defined as $\mu = \mathcal{N}(x \cap y) / \mathcal{N}(x)$, then for each entity mention in $x$, we check if it occurs in $y$, and if so, increment the intersection count  $\mathcal{N}(x \cap y)$ by unity. The second approach is followed in \cite{nan-etal-2021-entity}, In the first approach, we denote the metrics as $prec_s^{U}$, $prec_t^{U}$, $recall_t^{U}$, and $F1_t^{U}$ ($U$ indicates that only unique entity mentions are considered). In the second, we represent them as $prec_s^{NU}$, $prec_t^{NU}$, $recall_t^{NU}$, and $F1_t^{NU}$.

\section{Results}
\subsection{Comparison of the models}
In this sub-section, we report the results of the variations of fine-tuning the LED model on PubMed dataset. 
Table \ref{Table:par_all_types_rouge_meteor_bert} shows the F1-scores for ROUGE-1 (R-1), ROUGE-2 (R-2), ROUGE-L (R-L), BERTScore, and METEOR metrics along with values of the entity-level factual consistency metrics $prec_s^{U}$, $prec_s^{NU}$, $F1_t^{U}$  and $F1_t^{NU}$  on the filtered test dataset.
\begin{table*}[!htbp]
    \centering
    \begin{adjustbox}{width=1.0\linewidth}
    {\begin{tabular}{|p{3cm}cccccccccc|}  \hline
Model Name & R-1 & R-2 & R-L &R-LSum & METEOR & BERTScore  & $prec_s^{U}$ & $prec_s^{NU}$ & $F1_t^{U}$  & $F1_t^{NU}$ \\\hline
Fine-tuned LED &\bf{35.12} &\bf{14} & \bf{21.57} &\bf{29.96} &\bf{32.08} &\bf{84.96} &93.38  &94.76 &\bf{43.76}  &\bf{46.14}\\ \hline
Fine-tuned LED + Filtered dataset &33.18 &12.04 &19.93 &28.48 &27.43 &84.74 &\bf{96.04} &\bf{96.83} &40.15  &43.27\\ \hline
Fine-tuned LED + Filtered dataset + JAENS &30.21 &09.13 &18.26 &25.87 &23.55 &84.35 &92.16 &89.36 &40.15  &36.34 \\ \hline
\end{tabular} 
}
\end{adjustbox}
\caption{Evaluation of the models: F1-scores for ROUGE, METEOR, BERTScore, along with the $prec_s^{U}$, $prec_s^{NU}$, $F1_t^{U}$  and $F1_t^{NU}$ scores are used for evaluating the factual consistency of the generated summaries for the PubMed dataset. All scores in percentage (\%).} \label{Table:par_all_types_rouge_meteor_bert}
\end{table*}
The LED model fine-tuned on the filtered dataset achieves the highest $prec_s$ scores. However, when fine-tuning with LED is done without additional techniques like filtering or JEANS, the values of ROUGE, METEOR, BERTScore, and even F1$_t$ are the highest. This shows that not only $n$-gram matches and cosine similarity of embeddings are higher for the plain LED model, but the entity-level hallucination is also lower for it. Nan et al. \cite{nan-etal-2021-entity} also observed a reduction in ROUGE scores when data filtering and JEANS were applied, and remarked that it could be due to the increased complexity during decoding. Surprisingly, we find that data filtering and JEANS do not improve the $F1_t$ scores. In future, we intend to conduct a detailed study of this behaviour and try to decipher its reason. This could be related to the inaccuracy in entity recognition that we observed for the dataset; for example, on manual review, we found that many phrases detected as entities do not appear to be very important, but their match/mismatch between the generated and golden summary do impact the $F1_t$ scores. In contrast, in \cite{nan-etal-2021-entity}, standard entities are detected which could be achieved with high accuracy. Another difference with \cite{nan-etal-2021-entity} is that in our case, the dataset is much smaller and the summaries longer. 

%This indicates that the quality of the training dataset plays a vital role in reducing the number of hallucinations present in the generated summary. Adding the JAENS method helps to learn the identification of summary-worthy named-entities, and these important entities are incorporated in the summary through decoder self-attention. We also note that although filtering the dataset and incorporating the JAENS approach have proven effective in reducing hallucinated entities, they do have some impact on the ROUGE, METEOR, and BERTScore values. In fact, we can see that after applying the JAENS approach, these scores have actually decreased. Nan et al. \cite{nan-etal-2021-entity} made a similar observation for ROUGE scores and remarked that it could be due to the increased complexity during decoding.}
\subsection{Case study}
Figure \ref{fig:sample_Abs_RHS} shows sample outputs generated by the fine-tuning of the LED model without the filtered dataset, with the filtered dataset, and with both the filtered dataset and JAENS. The entities detected in each summary are also shown. The original abstract consists of 5 sentences, but after using entity-based filtering technique it consists of only 2 sentences.
In this case study, \colorbox{backgG}{yellow} color represents an incorrect representation of entity during summary generation. In this case study, \colorbox{backgPo}{cyan} color denotes an correct entity mention that was incorrectly generated by the fine-trained LED model. %While the model (Fine-tuned LED + Filtered dataset + JAENS) has eliminated the hallucinations produced by the other models, it does introduce a few new errors. Therefore, more research is needed to make these models practically viable.
 
\begin{figure*}[!htpb] 
\centering 
\small
\begin{tabular}{ |p{14cm}|} \hline
{\bf Ground-truth summary (abstract):} ``in recent years, evidence has accumulated that a dedicated plant cell wall integrity (cwi) maintenance mechanism exists. 
the available data suggest that it may represent a component of the stress response mechanisms underlying biotic and abiotic stress responses, which has not been identified previously as a distinct mechanism."\\
\textit{entities=`years', `evidence', `plant', `cell wall', `integrity', `cwi', `maintenance', `mechanism', `data', `stress response', `mechanisms', `biotic', `abiotic stress responses'}\\\hline
{\bf Fine-tuned LED:} ``plant cells need to maintain the functional integrity of their walls during cell morphogenesis and exposure to biotic / abiotic stress. 
the available evidence suggests that a dedicated plant cell wall integrity ( cwi ) maintenance mechanism exists. while our understanding of the mechanisms regulating stress responses and morphogenesis has increased significantly, our knowledge regarding the processes maintaining cwi is still limited. in the last years, a number of yeast strains have been identified and the available knowledge regarding cwi maintenance during plant development and in yeast has increased substantially. in this review, we will focus on the role of particular cell wall defects in the perception of and response to \colorbox{backgG}{botic stress responses}.\\
materials and \colorbox{backgG}{methodsa} yeast cwi monitoring and maintenance network is quite complex, providing an indication of the possible complexity of the interaction and the direct impact of these differences on the biological role and function of the \colorbox{backgG}{plant cwi.}"\\
\textit{entities=`plant cells', `functional', `integrity', `walls', `cell morphogenesis', `exposure to', `biotic', `abiotic stress', `evidence', `plant', `cell wall', `cwi', `maintenance', `mechanism', `mechanisms', `stress responses', `morphogenesis', `increased', `knowledge', `processes', `years', `yeast', `strains', `identified', `plant development', `review', `defects', `perception', `response', `botic stress', `methodsa', `monitoring', `network', `quite complex', `indication', `complexity', `interaction', `impact', `biological role', `function', `plant cwi'}\\\hline  
{\bf Fine-tuned LED + Filtered dataset:} `` plant cells need to maintain the functional integrity of their walls during cell morphogenesis and exposure to biotic / abiotic stress responses. 
the available evidence suggests that a dedicated plant cell wall integrity ( cwi ) maintenance mechanism exists. while the yeast cwi monitoring and maintenance network is quite complex, certain important differences exist that affect the biological role and function of the \colorbox{backgPo}{plant cwi maintenance mechanism}. in this review, we will focus on the effects of particular cell wall defects on pathogen resistance and the mode of action of specific signaling cascades."\\
\textit{entities=`plant cells', `functional', `integrity', `walls', `cell morphogenesis', `exposure to', `biotic', `abiotic stress responses', `evidence', `plant', `cell wall', `cwi', `maintenance', `mechanism', `yeast', `monitoring', `network', `quite complex', `biological role', `function', `review', `effects', `defects', `pathogen resistance', `mode', `signaling cascades'}\\\hline 
{\bf Fine-tuned LED + Filtered dataset + JAENS:} ``plant cells need to maintain the functional integrity of their walls during cell morphogenesis and exposure to biotic / abio stress responses. the available evidence suggests that a dedicated plant cwi maintenance mechanism exists, but what about the role of dedicated \colorbox{backgPo}{cells wall integrity ( cwi ) maintenance} during process."\\
\textit{entities=`plant cells', `functional', `integrity', `walls', `cell morphogenesis', `exposure to', `biotic', `abio stress responses', `evidence', `plant', `maintenance', `mechanism', `cells', `cwi', `process'}\\ \hline
\end{tabular} 	
\caption{Comparison of ground-truth and generated summary from the test set of the PubMed dataset. Hallucinations are highlighted. Input and ground-truth summary (abstract) taken from \texttt{\url{https://pubmed.ncbi.nlm.nih.gov/22629279/}}}
\label{fig:sample_Abs_RHS} 
\end{figure*}
\section{Conclusion}
We applied the Longformer Encoder-Decoder model on scientific research papers to generate summaries and used data filtering along with the JAENS approach to reduce entity hallucinations. We found that the simple fine-tuned LED model performs the best in terms of ROUGE, METEOR, and BERTScore but entity-based data filtering improves the scores of some of the factual consistency metrics. In the future, we would like to investigate in detail the reason behind the low performance of the JEANS approach. We also noticed that entities are not always identified with high recall and precision in the summary. We would like to analyze this issue in detail and improve the entity recognition module. Finally, we would like to study the reduction in ROUGE, METEOR, and BERTScore values that we observed in all the hallucination-mitigating designs.

%\clearpage
%\FloatBarrier
%\bibliographystyle{apacite}
\bibliographystyle{tfq}
\bibliography{anthology,custom}

\end{document}